\documentclass[letterpaper, 10 pt, conference]{ieeeconf}  

\IEEEoverridecommandlockouts                              

\usepackage{graphics} 
\usepackage{amsmath} 
\usepackage{amssymb}  

\usepackage{graphicx}
\usepackage{subcaption}
\usepackage{color}

\title{\LARGE \bf
Risk-Aware Path Planning for Ground Vehicles using Occluded Aerial Images
}

\author{Vishnu D. Sharma and Pratap Tokekar
\thanks{The authors are with the Department of Computer Science, University of Maryland College Park, USA 
        {\tt\small \{vishnuds, tokekar\}@umd.edu}. This work is supported by the National Science Foundation under Grant No. 1943368}%
}

\begin{document}

\maketitle
\thispagestyle{empty}
\pagestyle{empty}

\begin{abstract}
 We consider scenarios where a ground vehicle plans its path using data gathered by an aerial vehicle.
 In the aerial images, navigable areas of the scene may be occluded due to obstacles. Naively planning paths using aerial images may result in longer paths as a conservative planner may try to avoid regions that are occluded. We propose a modular, deep learning-based framework that allows the robot to \emph{predict} the existence of navigable areas in the occluded regions. Specifically, we use image inpainting methods over semantically segmented images to fill in parts of the areas that are potentially occluded, using a supervised neural network. However, the predictions from the neural network may not be accurate, introducing risk in navigating through the inpainted areas. Therefore, we extract uncertainty in these predictions and use a risk-aware approach that takes these uncertainties into account for path planning. We compare modules in our approach with traditional approaches to show the efficacy of the proposed framework through photo-realistic simulations. The modular pipeline allows further improvement in path planning and deployment in different settings. 
\end{abstract}

\section{INTRODUCTION}
\begin{figure}
      \centering
      \includegraphics[scale=0.7]{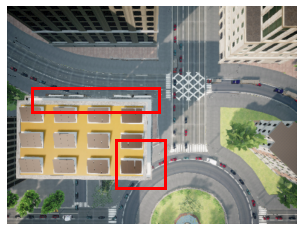}
      \caption{Examples of occlusions (shown in red rectangles) in an aerial image.}
      \label{fig:problem}
\end{figure}

Navigation in unknown areas in scenarios such as disaster response and post-accident scene analysis can be difficult and risky for in-field human supervision. To reduce the threat to human life, a team of an Unmanned Ground Vehicle (UGV) and an Unmanned Aerial Vehicle (UAV) can be deployed to help with such tasks. In this setting, the UAV flies above the area, capturing overhead images, and prescribes a path for navigation to the UGV, similar to finding a route on vehicle mapping services. As the UAV is closer to the ground than the satellite, it can capture more details about the area. However, unlike satellite images, which generate images with near-orthographic projection, a UAV flies at a much lower altitude and captures images with perspective projection. The objects closer to the UAV, such as a building, may occlude the navigable areas like roads (Figure ~\ref{fig:problem}) in the resulting images, making it difficult to identify a good path for UGV navigation. Similar issues may be encountered when a bridge occludes the aerial view of roadways or riverways, or in a forest where the tree canopy occludes the trails and roads. 

We focus on the scenarios where the UAV acts as a scout and captures the images of the environment, which are used later for UGV navigation or localization. The images captured by the UAV may not contain the views that could provide information about occluded regions, limiting the use of such images for path planning. Similar problems arise when the UGV can't control the UAV to gather more information.

The occluded regions in such scenarios are undesired and efficient path planning requires the information about the underlying region. We propose a learning-based framework to infer these regions and use the predicted navigable regions for path planning. This approach would also eliminate the need of capturing images from multiple viewpoints. As learning-based methods may not perform well over previously unseen data, we include the prediction uncertainty in the  planning framework. Specifically, we use an automated approach to identify occluded areas of interest, inpaint these areas with a deep learning network, identify navigable regions using a semantic segmentation network over the updated image, and perform a risk-aware path prediction between two points on the image. 

Following are our main contributions to this work: 
\begin{enumerate}
    \item We propose a planning framework for aerial images with occlusion, consisting of an automated masking procedure, an image inpainting network to replace the masked region with the underlying scene, and a risk-aware path planner. 
    \item We compare the proposed masking and inpainting approaches with other traditional approaches to provide the rationale for these choices and summarize their pros and cons.
    \item We show the effect of user-defined risk-aversion levels on the planned paths with and without the occlusion removal and demonstrate the benefit of our framework.
\end{enumerate} 

Our experiments show that this framework can help with planning through occluded, but potentially navigable areas.

\section{RELATED WORK}
A team consisting of a UAV and UGV can coordinate together by harnessing their respective advantages to perform complex tasks like search operations in hazardous areas~\cite{christie2017radiation}\cite{akhloufi2021unmanned}, exploration~\cite{hood2017bird}\cite{ropero2019terra}, and persistent monitoring~\cite{maini2020visibility}\cite{wu2020cooperative}. In such settings, the UAV can capture a bird's eye view of a large area, moving faster than a UGV due to fewer obstacles at high altitudes. The UGV, on the other hand, can capture the scene details, interact with the environment, and carry heavy payloads. The images from UAV serve as a global map and help the UGV identify safe regions and efficient routes. This setup effectively extends the typical scenarios where an overhead view of the environment is assumed to be provided and allows the UGV to work in a new environment. As this team moves through the environment the UGV uses the images from the UAV to localize itself by cross-view matching~\cite{dixit2020evaluation} and to navigate using the semantically segmented aerial map~\cite{toubeh2019risk}. 

Search methods can be used to find a path over the map generated by the UAV, but the quality of the path depends on the information captured. To capture such information, a UAV can sweep the area with a change in altitude when high-resolution images are required~\cite{wu2020cooperative}. Such approaches may not be useful when the task duration is limited and an existing image database may be used. In such images, the occluded regions are treated as obstacles and cannot be used for navigation. Satellite images also suffer from occlusion due to the presence of clouds and shadows for which image inpainting has been proposed as a viable solution~\cite{maalouf2009bandelet}\cite{siravenha2011evaluating}. Hsu et al.~\cite{hsu2017high} propose an inpainting network for front-facing aerial images, which does not cover as much information as overhead images can. To the best of our knowledge, there is no work on inpainting for aerial overhead images with perspective projection, captured with a downward-facing camera. In this work, we solve this problem with a generative adversarial network (GAN)~\cite{goodfellow2014generative}, variations of which have been shown to perform very well for a variety of image inpainting  tasks~\cite{weng2022survey}. Specifically, we use a partial convolution-based GAN~\cite{liu2018image} as it can inpaint irregular-shaped holes from occlusion masking  in our setup.

After inpainting the occlusion in the image, path planning can be done by applying semantic segmentation over the modified image to identify the navigable regions. Since the outcomes are just predictions, a safe operation requires taking the prediction uncertainty into account while planning. Gal et al.~\cite{gal2016dropout}, Tian et al.~\cite{tian2020uno} and Loquercio et al.\cite{loquercio2020general} present methods to extract uncertainty from deep learning model predictions. Sharma et al.~\cite{sharma2020risk} present a risk-aware planning framework for aerial images using the dropout-based approach~\cite{gal2016dropout}. We use this framework in our planning module with both the semantic segmentation and inpainted images acting as the source of uncertainty.

\begin{figure*}
      \centering
      \includegraphics[scale=0.20]{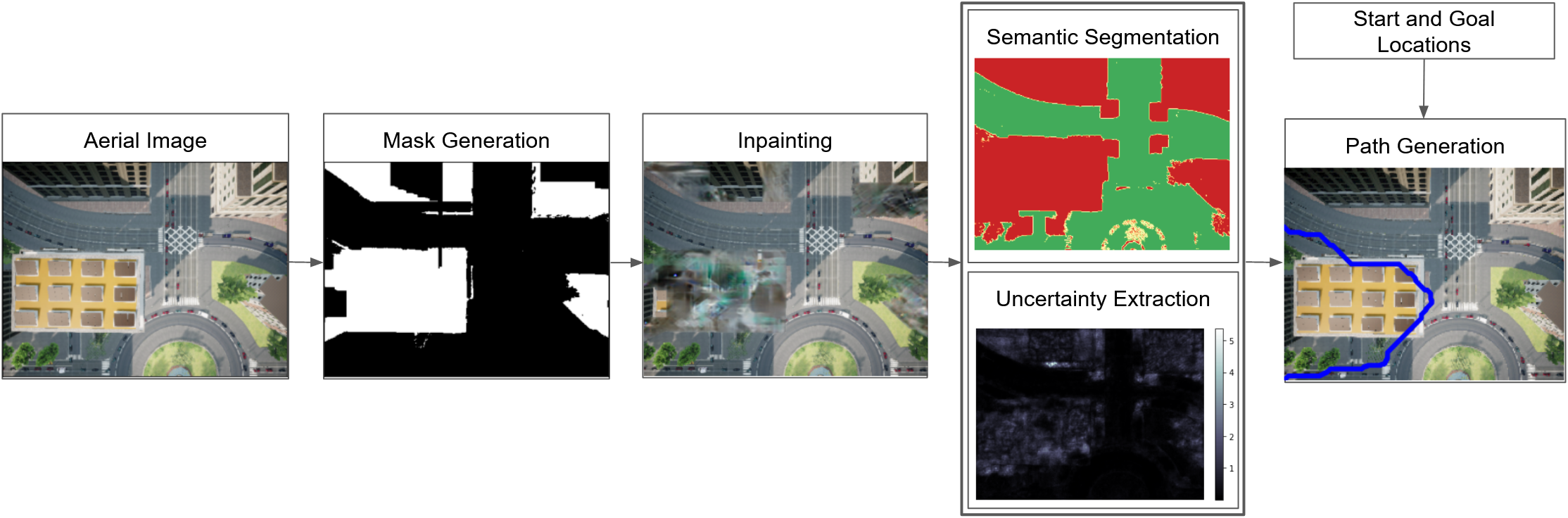}
      \caption{Overview of the proposed framework. Given an aerial image, a mask is generated indicating the parts that need to be modified. The image is inpainted using this mask, and then semantic segmentation and uncertainty map are generated. A path planner uses a combination of them to generate a risk-aware path. }
      \label{fig:overview}
\end{figure*} 

\section{METHOD}
Our approach comprises three key steps, as shown in Figure~\ref{fig:overview}: mask generation, inpainting, and path planning. We use the \textit{CityEnviron} environment in AirSim~\cite{shah2018airsim} simulator for our experiments. The following subsections detail each step of our approach.

\subsection{Mask Generation} 
\begin{figure}[h]
    \centering
    \begin{subfigure}[b]{0.20\textwidth}
        \centering
        \includegraphics[width=\textwidth]{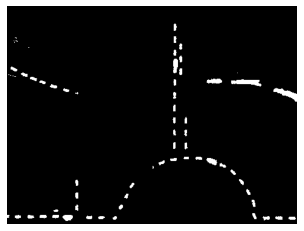}
        \caption[Road markers]%
        {{\small Road markers}}    
        \label{fig:road_marker}
    \end{subfigure}
    \hspace{0em}
    \begin{subfigure}[b]{0.20\textwidth}   
        \centering 
        \includegraphics[width=\textwidth]{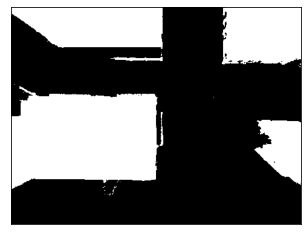}
        \caption[]%
        {{\small Buildings}}    
        \label{fig:buildings}
    \end{subfigure}
    \vspace{0em}
    \begin{subfigure}[b]{0.20\textwidth}  
        \centering 
        \includegraphics[width=\textwidth]{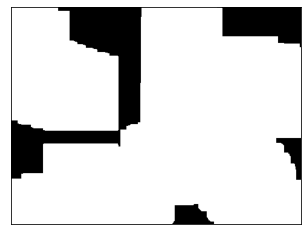}
        \caption[]%
        {{\small Dilated road markers}}    
        \label{fig:road_dilated}
    \end{subfigure}
    \hspace{0em}
    \begin{subfigure}[b]{0.20\textwidth}   
        \centering 
        \includegraphics[width=\textwidth]{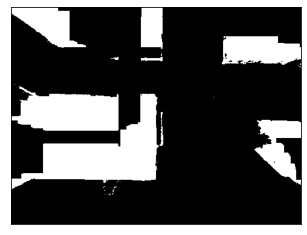}
        \caption[]%
        {{\small Final mask ($b \bigodot c$)}}    
        \label{fig:final_mask}
    \end{subfigure}
    \caption[ Stages of mask generation ]
    {\small Stages of mask generation.} 
    \label{fig:masking}
\end{figure}

Inpainting networks require a mask as an input to identify the areas to modify/inpaint over. In this work, we treat the buildings near the road as the obstacle to be removed and generate masks using them. Figure~\ref{fig:masking} shows the steps involved in this process. We first find the road markers ((Figure~\ref{fig:road_marker}) and buildings (Figure ~\ref{fig:buildings}) using Bayesian SegNet~\cite{kendall2016bayesian}, a semantic segmentation network, trained over aerial image segmentation data~\cite{sharma2020risk}. The road marker help in identifying the center of the road and the areas within the width of the road are marked by performing dilation over the binarized road marker image (Figure~\ref{fig:road_dilated}). We assume the objects in aerial images to have a rectilinear structure and thus use a square mask. The size of the dilation kernel depends on the UAV altitude. Lastly, we perform a bitwise AND operation between the resulting image and the binary image from buildings to obtain a minimal mask for inpainting the areas near the road occluded by the buildings (Figure~\ref{fig:final_mask}).


\subsection{Inpainting} The mask generated in the previous step may have irregularly shaped regions even with a square dilation kernel, thus we use a partial convolutions-based GAN~\cite{liu2018image}. Our training dataset is smaller compared to similar inpainting approaches and is virtually increased to some extent with randomly generated masks during training. We use the aerial image dataset from Dixit et al.~\cite{dixit2020evaluation}, which contains images unobstructed from \textit{CityEnviron} to make the network learn how to inpaint roads.

\subsection{Path Planning} To find the path between two locations on an aerial map, we perform semantic scene parsing with Bayesian SegNet and assign costs to different object classes. Additionally, we quantify the prediction uncertainty as the average variance in the predictions, i.e., $\texttt{Uncert}(l_x) = \frac{1}{K} \sum_k \texttt{Var}_n(P(l_x|X))$, where $l_x$ is label assigned to the pixel $x$,  $X$ is the image,  $K$ is the number of classes, and $\texttt{Var}_n(\cdot)$ denotes the variance calculated over $n$ samples. We use the risk-aware planning proposed by Sharma et al.~\cite{sharma2020risk} to find the path between two locations while considering the prediction uncertainty, using the cost matrix obtained as follows: 
\begin{align}
        \hat C(x) = C(l_x) + \lambda \cdot  \texttt{Uncert}(l_x), 
        \label{eq:costmat}
\end{align}
where $C(l_x)$ is the cost map denoting the user-defined cost assigned to the pixel $l_x$. $\lambda$ is a user-defined parameter to control the level risk-aversion in planner. A higher values of $\lambda$ results in a more conservative path.

\section{EXPERIMENTS AND RESULTS}
\begin{figure}
    \centering
    \begin{subfigure}{0.23\textwidth}
        \centering
        \includegraphics[width=\textwidth]{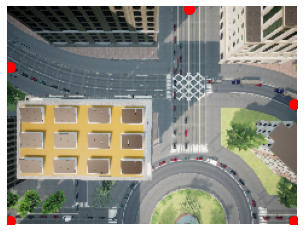}
        \caption[Perspective Scene]%
        {{\small Perspective Scene}}    
        \label{fig:base_scene}
    \end{subfigure}
    \hspace{0em}
    \begin{subfigure}{0.23\textwidth}   
        \centering 
        \includegraphics[width=\textwidth]{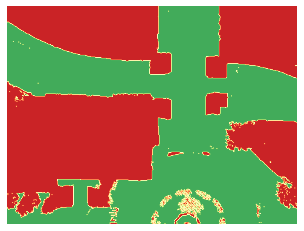}
        \caption[Perspective Segmentation]%
        {{\small Perspective Segmentation}}    
        \label{fig:base_seg}
    \end{subfigure}
    \label{fig:base_imgs}
    \vspace{0em}
    \begin{subfigure}{0.23\textwidth}  
        \centering 
        \includegraphics[width=\textwidth]{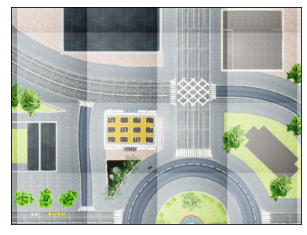}
        \caption[Orthographic Scene]%
        {{\small Orthographic Scene}}    
        \label{fig:gt_scene}
    \end{subfigure}
    \hspace{0em}
    \begin{subfigure}{0.23\textwidth}   
        \centering 
        \includegraphics[width=\textwidth]{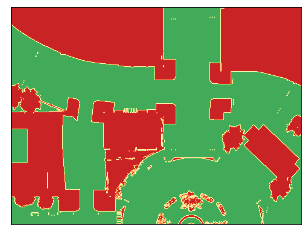}
        \caption[Orthographic Segmentation]%
        {{\small Orthographic Segmentation}}    
        \label{fig:gt_seg}
    \end{subfigure}
    \caption[ Reference input (perspective) and approximated ground truth (orthographic) images. The red dots show the locations used for calculating surprise. ]
    {\small  Reference input (perspective) and approximated ground truth (orthographic) images. The red dots show the locations used for calculating surprise.} 
    \label{fig:gt_imgs}
\end{figure}

We conduct experiments to justify the chosen methods for parts of our pipelines and then show the effect of using them together for path planning. We start by comparing our masking approach with a manual approach, then we compare the inpainting approach with different non-learning-based approaches, and finally show the effect of changing $\lambda$ on the prescribed path; some examples of these are visualized at our project website\footnote{http://raaslab.org/projects/occlusion-inpainting.html}. As AirSim does not allow removing objects like building from the scene in \textit{CityEnviron}, we can not directly compare the results for the first two and present the qualitative comparison only. Figure~\ref{fig:gt_imgs} shows the reference images, the orthographic projection obtained by moving the UAV to multiple locations and manually cropping and stacking the images and their semantic segmentations.

\begin{figure}
      \centering
      \includegraphics[scale=0.50]{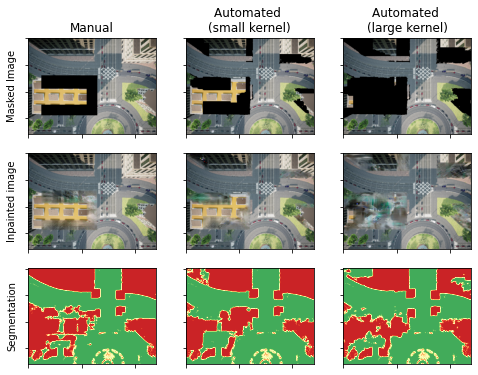}
      \caption{\small Comparison of non-learning and learning-based approaches for masking.}
      \label{fig:mask_exp}
\end{figure}

Our masking method relies on semantic segmentation for identifying the occlusions and selecting only the region in the vicinity of the road for inpainting.  We compare this approach with a manual selection of masking regions. Since the choice of dilation kernel affects the size of the regions marked for inpainting, we also show the effect of using a small kernel of size $121\times121$, and a large kernel of size $151\times151$. The resulting segmentation on the inpainted image is also presented, as shown in Figure~\ref{fig:mask_exp}. We observed that manual masking results in more regular and controlled inpainting. Automated making with a small kernel produces masks similar in form and segmentation to the manual mask, but the resulting regions may not be connected well when road markers are occluded. The larger kernel  masks all the edges of the building and also masks other occlusions in the scene (top and right parts of the image) better than the small kernel, making it more suitable when there are multiple occluded regions in the scene. Thus, we use the larger kernel in our experiments.

The training data for the inpainting network contains images captured at different heights that increase the amount and variety of training examples.
We compare it against three other approaches: (a) filling the masked areas with average values of pixels belonging to the road in the image (\textit{mean-replacement}), (b) replacing the masked areas with a road-like patch generated manually (\textit{patch-replacement}), and (c) using Navier-Stokes inpainting method~\cite{bertalmio2001navier} which uses ideas from fluid dynamics for inpainting (\textit{Navier-Stokes}).

\begin{figure}
      \centering
      \includegraphics[scale=0.45]{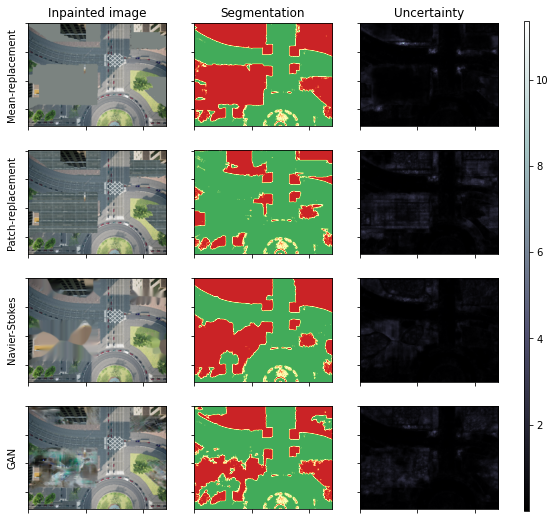}
      \caption{\small Comparison of approaches for inpainting.}
      \label{fig:inpt_exp}
\end{figure}

Figure~\ref{fig:inpt_exp} shows a comparison of the results for inpainting methods. Mean-replacement results in segmentation with the masked areas identified as the obstacles. It is almost similar to the input reference (Figure~\ref{fig:base_seg}) and thus does not improve planning. Patch-replacement makes most of the building navigable, as we are replacing the mask with pixels from a navigable class. This results in over-inpainting and can result in a collision with obstacles. Navier-Stokes results in very smooth inpainting, but the semantic maps have limited growth in the potential navigable region. GAN is more controlled and accurate compared to these methods. We also expect a good method to have higher uncertainty compared to the inpainted regions compared to the non-inpainted ones, as these are just predictions, not observations. Mean-replacement has a very low uncertainty in these areas, which further makes it unsuitable. GAN has a slightly higher uncertainty in the masked regions, which makes it more appropriate for the risk-aware path planning problem.

\begin{figure}
      \centering
      \includegraphics[scale=0.5]{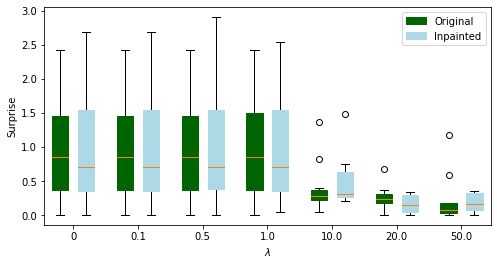}
      \caption{\small Surprise for various $\lambda$s with and without inpainting.}
      \label{fig:surprise}
\end{figure}

For path planning, we perform semantic segmentation over the inpainted image using Bayesian SegNet trained over  480 images, split into training and test sets in a ratio of 13:3. We use Eq.~\ref{eq:costmat} with a cost of 1 for navigable regions like roads and a cost of 10 for the rest. Variance is computed with $n=20$ runs. To compare the effect of varying $\lambda$, we define a quantity called \textit{surprise} as follows:
\begin{align}
    \texttt{surprise} = \frac{ \sum_{l_x \in path} C_{true}(l_x) - C_{pred}(l_x)}{length(path)},
\end{align}
where $path$ is the path over pixels as prescribed by the planner, and $C_{true}$ and $C_{pred}$ are the cost maps generated over semantic segmentation outputs from the ground truth (shown in Figure~\ref{fig:gt_seg}) and inpainted images, respectively. A higher surprise indicates that a larger part of the path passes through risky regions. We pick 5 locations on the scene image, as shown in Figure~\ref{fig:base_scene}, and compute surprise over different source-goal combinations. The results over the paths generated with and without our inpainting-aided approach for different values of $\lambda$ are summarized in Figure~\ref{fig:surprise}. We observed that the inpainted image results in slightly lesser surprise, on average,  than the original image, as inpainting allows planning paths through the occluded, but potentially navigable areas. Both approaches have a higher variance for smaller $\lambda$s as it encourages them to pass through uncertain regions for a shorter path. As we increase $\lambda$, the planner gets risk-averse and the path passes through the regions with a high chance of being navigable. In these cases, the surprise for both gets close to zero and has a much smaller variance.



\section{CONCLUSIONS}
In this work, we propose a modular, risk-aware path planning framework for aerial images containing occlusions. We use  image inpainting and semantic segmentation networks to identify occluded areas, predict the region beneath, and plan a risk-aware path considering the uncertainty in predictions. Our comparisons highlight the justification and advantages of the underlying modules. The modular design opens multiple avenues for future work, including improvements in individual modules, and combined networks for inpainted segmentation. With appropriate substitutions, the framework can also be used for street-view images for which many high-performance pre-trained detection and inpainting models are available. 



\bibliographystyle{IEEEtran}
\bibliography{bibfile}

\begin{thebibliography}{10}
\providecommand{\url}[1]{#1}
\csname url@rmstyle\endcsname
\providecommand{\newblock}{\relax}
\providecommand{\bibinfo}[2]{#2}
\providecommand\BIBentrySTDinterwordspacing{\spaceskip=0pt\relax}
\providecommand\BIBentryALTinterwordstretchfactor{4}
\providecommand\BIBentryALTinterwordspacing{\spaceskip=\fontdimen2\font plus
\BIBentryALTinterwordstretchfactor\fontdimen3\font minus
  \fontdimen4\font\relax}
\providecommand\BIBforeignlanguage[2]{{%
\expandafter\ifx\csname l@#1\endcsname\relax
\typeout{** WARNING: IEEEtran.bst: No hyphenation pattern has been}%
\typeout{** loaded for the language `#1'. Using the pattern for}%
\typeout{** the default language instead.}%
\else
\language=\csname l@#1\endcsname
\fi
#2}}

\bibitem{christie2017radiation}
G.~Christie, A.~Shoemaker, K.~Kochersberger, P.~Tokekar, L.~McLean, and
  A.~Leonessa, ``Radiation search operations using scene understanding with
  autonomous uav and ugv,'' \emph{Journal of Field Robotics}, vol.~34, no.~8,
  pp. 1450--1468, 2017.

\bibitem{akhloufi2021unmanned}
M.~A. Akhloufi, A.~Couturier, and N.~A. Castro, ``Unmanned aerial vehicles for
  wildland fires: Sensing, perception, cooperation and assistance,''
  \emph{Drones}, vol.~5, no.~1, p.~15, 2021.

\bibitem{hood2017bird}
S.~Hood, K.~Benson, P.~Hamod, D.~Madison, J.~M. O'Kane, and I.~Rekleitis,
  ``Bird's eye view: Cooperative exploration by ugv and uav,'' in \emph{2017
  International Conference on Unmanned Aircraft Systems (ICUAS)}.\hskip 1em
  plus 0.5em minus 0.4em\relax IEEE, 2017, pp. 247--255.

\bibitem{ropero2019terra}
F.~Ropero, P.~Mu{\~n}oz, and M.~D. R-Moreno, ``Terra: A path planning algorithm
  for cooperative ugv--uav exploration,'' \emph{Engineering Applications of
  Artificial Intelligence}, vol.~78, pp. 260--272, 2019.

\bibitem{maini2020visibility}
P.~Maini, P.~Tokekar, and P.~B. Sujit, ``Visibility-based persistent monitoring
  of piece-wise linear features on a terrain using multiple aerial and ground
  robots,'' \emph{{IEEE} Transactions on Automation Science and Engineering},
  pp. 1--13, 2020, conditionally accepted.

\bibitem{wu2020cooperative}
Y.~Wu, S.~Wu, and X.~Hu, ``Cooperative path planning of uavs \& ugvs for a
  persistent surveillance task in urban environments,'' \emph{IEEE Internet of
  Things Journal}, vol.~8, no.~6, pp. 4906--4919, 2020.

\bibitem{dixit2020evaluation}
D.~Dixit and P.~Tokekar, ``Evaluation of cross-view matching to improve ground
  vehicle localization with aerial perception,'' \emph{arXiv preprint
  arXiv:2003.06515}, 2020.

\bibitem{toubeh2019risk}
M.~Toubeh and P.~Tokekar, ``Risk-aware planning by confidence estimation using
  deep learning-based perception,'' \emph{arXiv preprint arXiv:1910.00101},
  2019.

\bibitem{maalouf2009bandelet}
A.~Maalouf, P.~Carr{\'e}, B.~Augereau, and C.~Fernandez-Maloigne, ``A
  bandelet-based inpainting technique for clouds removal from remotely sensed
  images,'' \emph{IEEE transactions on geoscience and remote sensing}, vol.~47,
  no.~7, pp. 2363--2371, 2009.

\bibitem{siravenha2011evaluating}
A.~C. Siravenha, D.~Sousa, A.~Bispo, and E.~Pelaes, ``Evaluating inpainting
  methods to the satellite images clouds and shadows removing,'' in
  \emph{International Conference on Signal Processing, Image Processing, and
  Pattern Recognition}.\hskip 1em plus 0.5em minus 0.4em\relax Springer, 2011,
  pp. 56--65.

\bibitem{hsu2017high}
C.~Hsu, F.~Chen, and G.~Wang, ``High-resolution image inpainting through
  multiple deep networks,'' in \emph{2017 International Conference on Vision,
  Image and Signal Processing (ICVISP)}.\hskip 1em plus 0.5em minus 0.4em\relax
  IEEE, 2017, pp. 76--81.

\bibitem{goodfellow2014generative}
I.~J. Goodfellow, J.~Pouget-Abadie, M.~Mirza, B.~Xu, D.~Warde-Farley, S.~Ozair,
  A.~Courville, and Y.~Bengio, ``Generative adversarial networks,'' \emph{arXiv
  preprint arXiv:1406.2661}, 2014.

\bibitem{weng2022survey}
Y.~Weng, S.~Ding, and T.~Zhou, ``A survey on improved gan based image
  inpainting,'' in \emph{2022 2nd International Conference on Consumer
  Electronics and Computer Engineering (ICCECE)}.\hskip 1em plus 0.5em minus
  0.4em\relax IEEE, 2022, pp. 319--322.

\bibitem{liu2018image}
G.~Liu, F.~A. Reda, K.~J. Shih, T.-C. Wang, A.~Tao, and B.~Catanzaro, ``Image
  inpainting for irregular holes using partial convolutions,'' in
  \emph{Proceedings of the European Conference on Computer Vision (ECCV)},
  2018, pp. 85--100.

\bibitem{gal2016dropout}
Y.~Gal and Z.~Ghahramani, ``Dropout as a bayesian approximation: Representing
  model uncertainty in deep learning,'' in \emph{international conference on
  machine learning}.\hskip 1em plus 0.5em minus 0.4em\relax PMLR, 2016, pp.
  1050--1059.

\bibitem{tian2020uno}
J.~Tian, W.~Cheung, N.~Glaser, Y.-C. Liu, and Z.~Kira, ``Uno: Uncertainty-aware
  noisy-or multimodal fusion for unanticipated input degradation,'' in
  \emph{2020 IEEE International Conference on Robotics and Automation
  (ICRA)}.\hskip 1em plus 0.5em minus 0.4em\relax IEEE, 2020, pp. 5716--5723.

\bibitem{loquercio2020general}
A.~Loquercio, M.~Segu, and D.~Scaramuzza, ``A general framework for uncertainty
  estimation in deep learning,'' \emph{IEEE Robotics and Automation Letters},
  vol.~5, no.~2, pp. 3153--3160, 2020.

\bibitem{sharma2020risk}
V.~Sharma, M.~Toubeh, L.~Zhou, and P.~Tokekar, ``Risk-aware planning and
  assignment for ground vehicles using uncertain perception from aerial
  vehicles,'' in \emph{Proceedings of the IEEE/RSJ International Conference on
  Intelligent Robots and Systems (IROS)}, 2020.

\bibitem{shah2018airsim}
S.~Shah, D.~Dey, C.~Lovett, and A.~Kapoor, ``Airsim: High-fidelity visual and
  physical simulation for autonomous vehicles,'' in \emph{Field and service
  robotics}.\hskip 1em plus 0.5em minus 0.4em\relax Springer, 2018, pp.
  621--635.

\bibitem{kendall2016bayesian}
A.~Kendall, V.~Badrinarayanan, and R.~Cipolla, ``Bayesian segnet: Model
  uncertainty in deep convolutional encoder-decoder architectures for scene
  understanding,'' 2016.

\bibitem{bertalmio2001navier}
M.~Bertalmio, A.~L. Bertozzi, and G.~Sapiro, ``Navier-stokes, fluid dynamics,
  and image and video inpainting,'' in \emph{Proceedings of the 2001 IEEE
  Computer Society Conference on Computer Vision and Pattern Recognition. CVPR
  2001}, vol.~1.\hskip 1em plus 0.5em minus 0.4em\relax IEEE, 2001, pp. I--I.

\end{thebibliography}

\end{document}